\documentclass[letterpaper, 10 pt, conference]{style/ieeeconf}  % Comment this line out
\IEEEoverridecommandlockouts                              % This command is only
\date{}
\usepackage[english]{babel}
\usepackage{mathtools} % loads amsmath 
\usepackage{graphicx}
\usepackage{wrapfig}
\usepackage{multicol}
\usepackage{subfig}
\usepackage{booktabs}
\usepackage{amsfonts}
\usepackage{gensymb}
\usepackage{listings}
\usepackage{pifont}

\usepackage{lipsum}                     % Dummytext
\usepackage{xargs}                      % Use more than one optional parameter in a new commands
\usepackage[pdftex,dvipsnames]{xcolor}  % Coloured text etc.
\usepackage[colorinlistoftodos,prependcaption,textsize=tiny]{todonotes}
\newcommandx{\unsure}[2][1=]{\todo[linecolor=red,backgroundcolor=red!25,bordercolor=red,#1]{#2}}
\newcommandx{\change}[2][1=]{\todo[linecolor=blue,backgroundcolor=blue!25,bordercolor=blue,#1]{#2}}
\newcommandx{\info}[2][1=]{\todo[linecolor=OliveGreen,backgroundcolor=OliveGreen!25,bordercolor=OliveGreen,#1]{#2}}
\newcommandx{\improvement}[2][1=]{\todo[linecolor=Plum,backgroundcolor=Plum!25,bordercolor=Plum,#1]{#2}}
\newcommandx{\thiswillnotshow}[2][1=]{\todo[disable,#1]{#2}}
\usepackage[colorinlistoftodos,prependcaption,textsize=tiny]{todonotes}

\usepackage[normalem]{ulem}

\newcommand{\matrixdim}[2]{\underset{\scriptscriptstyle#2}{#1}}

\newcommand{\ra}[1]{\renewcommand{\arraystretch}{#1}}

\usepackage[bottom]{footmisc}
\usepackage[colorlinks=true, allcolors=blue]{hyperref}

\usepackage{array,tabularx,calc}

\newenvironment{conditions}[1][where:]
  {%
   #1\tabularx{\textwidth-\widthof{#1}}[t]{
     >{$}l<{$} @{${}={}$} X@{}
   }%
  }
  {\endtabularx\\[\belowdisplayskip]}

\newcommand{\xmark}{\text{\ding{55}}}
\newcommand{\cmark}{\text{\ding{51}}}

\usepackage{makecell}
\usepackage{colortbl}
\definecolor{light-gray}{gray}{0.9}
\newcolumntype{C}{>{\columncolor{light-gray}}c}
\newcolumntype{L}{>{\columncolor{light-gray}}l}
\newcolumntype{R}{>{\columncolor{light-gray}}r}

\usepackage{filecontents}

\title{\LARGE \bf
The CoSTAR Block Stacking Dataset:\\Learning with Workspace Constraints
}

\author{Andrew Hundt$^1$, Varun Jain$^1$, Chia-Hung Lin$^1$, Chris Paxton$^{2}$, Gregory D. Hager$^1$
\thanks{
$^{1}$ Johns Hopkins University Department of Computer Science. \{ahundt, vjain, ch.lin, cpaxton, ghager1\}@jhu.edu}
\thanks{$^{2}$ Chris Paxton is with NVIDIA, USA}}

\begin{document}
\maketitle
\begin{abstract} 
A robot can now grasp an object more effectively than ever before, but once it has the object what happens next?
We show that a mild relaxation of the task and workspace constraints implicit in existing object grasping datasets can cause neural network based grasping algorithms to fail on even a simple block stacking task when executed under more realistic circumstances.

To address this, we introduce the JHU CoSTAR Block Stacking Dataset (BSD), where a robot interacts with 5.1 cm colored blocks to complete an order-fulfillment style block stacking task. 
It contains dynamic scenes and real time-series data in a less constrained environment than comparable datasets.
There are nearly 12,000 stacking attempts and over 2 million frames of real data.
We discuss the ways in which this dataset provides a valuable resource for a broad range of other topics of investigation.

We find that hand-designed neural networks that work on prior datasets do not generalize to this task. Thus, to establish a baseline for this dataset, we demonstrate an automated search of neural network based models using a novel multiple-input HyperTree MetaModel, and find a final model which makes reasonable 3D pose predictions for grasping and stacking on our dataset.

The CoSTAR BSD, code, and instructions are available at 
\url{sites.google.com/site/costardataset}.
% \url{github.com/jhu-lcsr/costar_plan}.
\end{abstract}

\section{Introduction}

Existing task and motion planning algorithms are more than robust enough for a wide variety of impressive tasks, and the community is looking into environments that are ever closer to truly unstructured scenes. 
% However, as we seek to move towards more unstructured environments we would like to learn to work within unstructured scenes to complete tasks we have never modeled before. 
In this context, the recent success of Deep Learning (DL) on challenging computer vision tasks has spurred efforts to develop DL systems that can be applied to perception-based robotics~\cite{levine2016learning,mahler2017dex}.
DL promises end-to-end training from representative data, to solve complex, perception-based robotics tasks in realistic environments with higher reliability and less programming effort than traditional programming methods. 
% As new methods are developed, we will need to empirically evaluate the quality of model based algorithms against their unstructured peers. 
Data from existing planning methods can provide an excellent source of ground truth data against which we can evaluate new methods and compare the quality of model based algorithms against their unstructured peers. 
% Data from the high quality planning methods we know and love can provide an excellent source of ground truth data against which we can evaluate new methods and compare the quality of model based algorithms against their unstructured peers.

\begin{figure}[bt!]
    \centering
    \hfill
    \includegraphics[width=\columnwidth]{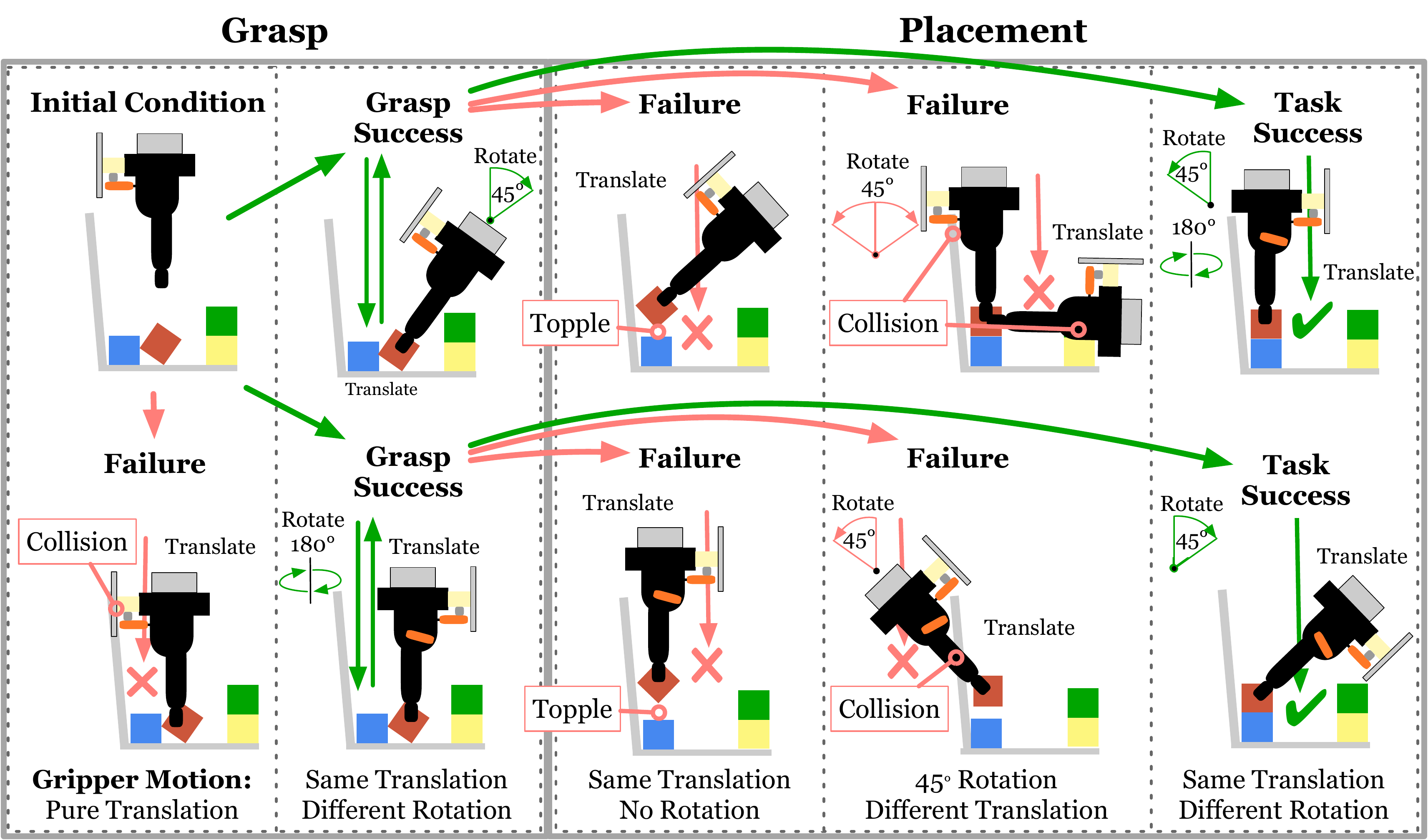}
    \caption{
    \label{fig:gripper_asymmetry_time_dependencies}
    A simplified 2-step  \texttt{grasp(red)}, \texttt{place(red, on\_blue)} stacking task with a side wall and asymmetric gripper. Arrows indicate possible sequences of actions. Task success is affected by the shape of the gripper, the obstacles, and the relationship between the pose of the gripper and the block it is holding.
}
\vspace{-0.2cm}
\end{figure}

\begin{figure}[bt!]
    \centering
    \hfill
    \includegraphics[width=\columnwidth]{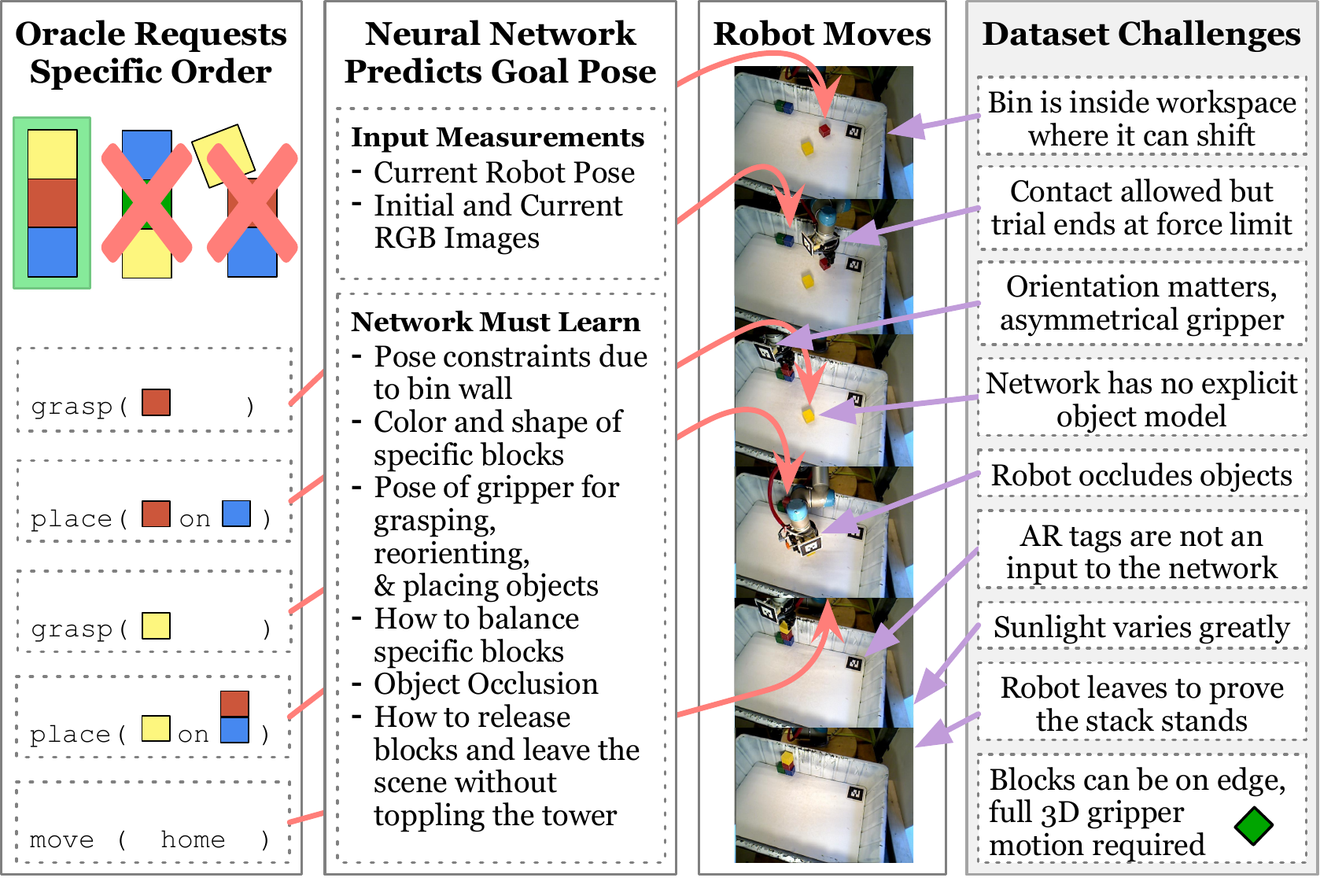}
    \caption{
    \label{fig:StackingSkillsNeuralNetwork} %Task Action 
    An overview of the CoSTAR Block Stacking Dataset task and the requirements placed on our example neural network. 
    In each example of stacking, the oracle requests a random specific order of colored blocks to simulate different customer choices.
}
\vspace{-0.4cm}
\end{figure}

\begin{figure}[btp!]
\centering
\includegraphics[width=0.75\columnwidth]{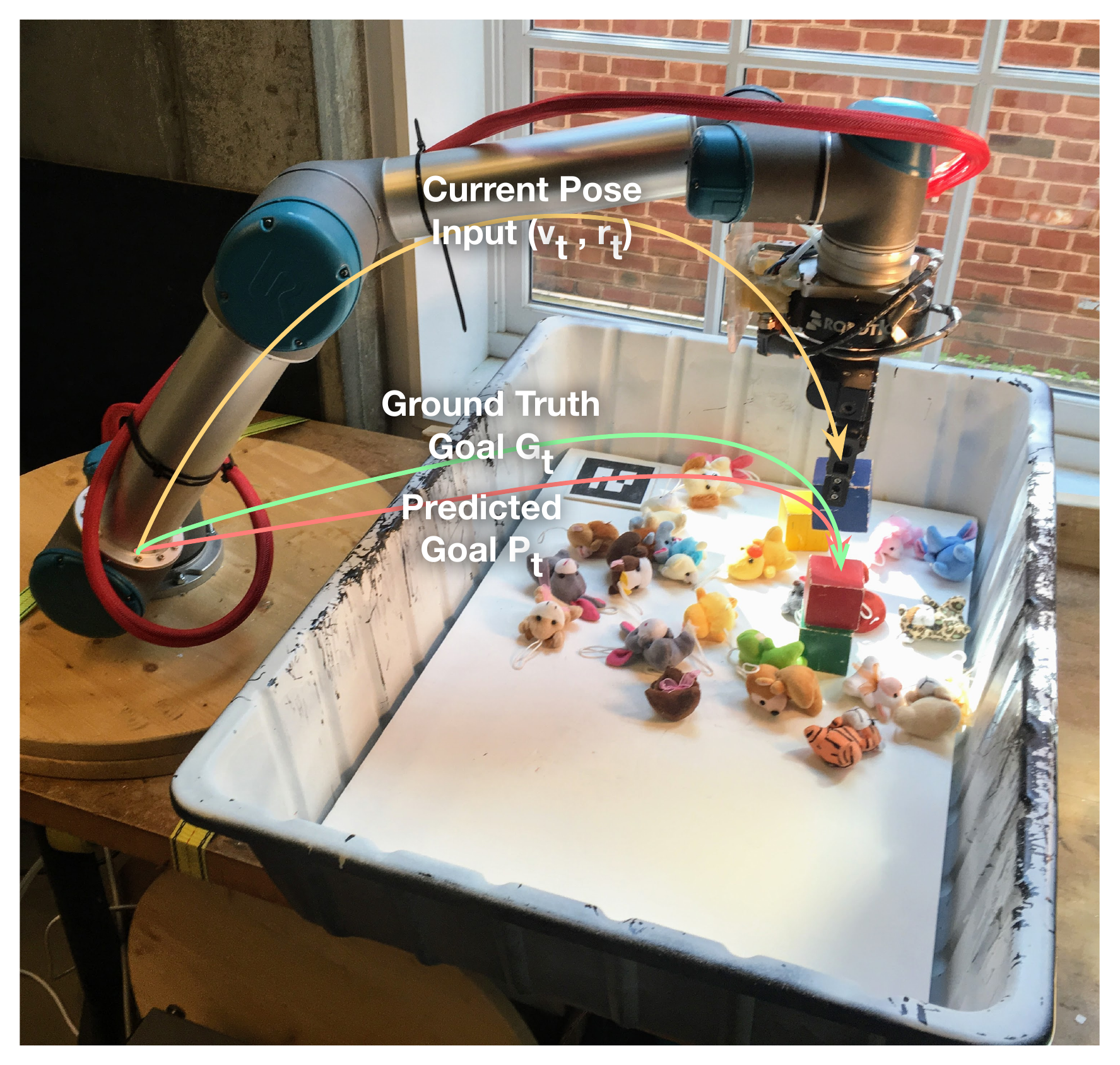}
%\vfill
% \includegraphics[width=0.4\columnwidth]{workspace}
% \vfill
% \includegraphics[width=\columnwidth]{ur5_simulation.png}
\caption{The CoSTAR system~\cite{paxton2017costar} collecting the block stacking dataset. 
% Each example consists of an attempt to stack 3 colored blocks in a specified order. 
% The input current pose and output predicted pose $P_t$ of the robot consist of rotation $r$ and translation $v$ of the gripper relative to the robot base frame, which are encoded for the HyperTree neural network as detailed in equations \ref{eq:translation}, \ref{eq:rotation}, \ref{eq:example}, and \ref{eq:goal}.
% TODO(ahundt) add base, ee_link, and gripper_center frames to picture
}
\label{fig:workspace}
\vspace{-0.2cm}
\end{figure}

\begin{figure*}[btp!]
\centering
% \includegraphics[width=0.9\textwidth]{2018-05-15-10-45-10_example000006_success_tiled.jpg}
% \vspace{10pt}
\includegraphics[width=0.9\textwidth]{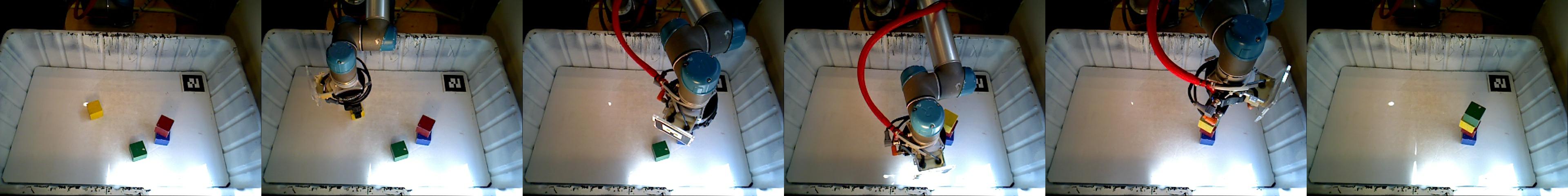}
% \vspace{0.1cm}
% \includegraphics[width=0.9\textwidth]{2018-07-05-11-45-26_example000005_success_tiled.jpg}
% % \vspace{0.1cm}
% % \includegraphics[width=2.0\columnwidth]{2018-08-02-19-36-36_example000003_success_tiled.jpg}
% \includegraphics[width=0.9\textwidth]{2018-08-06-15-45-17_example000002_success_tiled.jpg}
% % \vspace{0.1cm}
\includegraphics[width=0.9\textwidth]{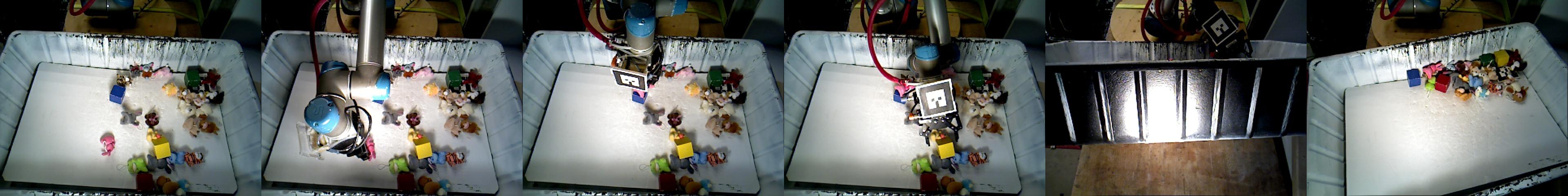}

\caption{ Row 1 is a successful and row 2 is a failed block stacking attempt. A sequence starts on the left with a clear view at frame $I_0$ then proceeds right showing the timesteps of the 5 goal poses $G_t$ (Eq. \ref{eq:goal}, Fig. \ref{fig:StackingSkillsNeuralNetwork}, \ref{fig:workspace}) at which the gripper may open or close. Notice the variation in bin position, gripper tilt, the challenging lighting conditions, the stack of 4 blocks, and the object wear. Viewing video and other details is highly recommended, see \url{sites.google.com/site/costardataset}.
% \ref{fig:MotionNetwork}.
}
\label{fig:attempt}
\vspace{-0.4cm}
\end{figure*}

Existing robotics datasets such as those outlined in  Table \ref{table:datasetcomparison} provide a good representation of certain aspects of manipulation,  but fail to capture end-to-end task planning with obstacle avoidance.
% Unfortunately, this means the often milimeter-sized boundary between success and failure is a critical gap we cannot yet asses. 
Capturing the interaction between the robot, objects, and obstacles is critical to ensure success in dynamic environments, as we show in Fig. \ref{fig:gripper_asymmetry_time_dependencies}.
How can we investigate these dependencies within a dynamic scene? 
Can an implicit understanding of physical dependencies be created from raw data?
We introduce the CoSTAR Block Stacking Dataset (Fig. \ref{fig:StackingSkillsNeuralNetwork}, \ref{fig:workspace}, \ref{fig:attempt}, and Sec. \ref{ssec:datacollection}) for the purpose of investigating these questions. 
It is designed as a benchmark for performing complex, multi-step manipulation tasks in challenging scenes. 
The target task is stacking 3 of 4 colored blocks in a specific order with simple target objects in a cluttered scene and variable surrounding environment.

Our block stacking task is constrained enough that one dataset might cover the task sufficiently, while still ensuring dynamics and physical dependencies are part of the environment.
We show how, despite this simplicity, the task cannot be completed with the current design of existing grasping networks (Sec. \ref{ssec:datacollection}), nor by the trivial transfer of one example underlying architecture to a 3D control scheme (Sec. \ref{sec:problemandapproach}).
Therefore, we apply Neural Architecture Search (NAS)\cite{2018nassurvey} to this dataset using our novel multiple-input HyperTree MetaModel (Fig. \ref{fig:MotionNetwork} and Sec. \ref{ssec:hypertreeMetaModel}) to find a viable model.  
NAS is an approach to automatically optimize neural network based models to specific applications.
In fact, we show that useful training progress is made with only a small subset of network models from across a broad selection of similar architectures (Fig \ref{fig:HyperTreeModelSearchComparisonOfError}). 
We hope that with specialization to other particular tasks, MetaModels based on HyperTrees might also serve to optimize other applications which incorporate multiple input data sources.

To summarize, we make the following contributions:
\begin{enumerate}
\item The CoSTAR Block Stacking Dataset: a valuable resource to researchers across a broad range of robotics and perception investigations.
\item The HyperTree MetaModel, which describes a space for automatically refining neural network models with multiple input data streams. 
\item Baseline architectures to predict 6 Degree of Freedom (DOF) end-effector goals for the grasping and placement of specific objects, as found via HyperTree search.
\end{enumerate}

\begin{table}\centering
\begin{tabular}{llr}
\toprule
\multicolumn{3}{c}{CoSTAR Block Stacking Dataset Summary} \\
\cmidrule(r){1-3}
\textbf{Calibrated Images} & \multicolumn{2}{l}{color, depth}       \\
\textbf{Joint Data} & \multicolumn{2}{l}{angle, velocity}\\
\textbf{Labels}       & \multicolumn{2}{l}{action, success/failure/error.failure}    \\
\textbf{Blocks}       & \multicolumn{2}{l}{red, green, yellow, blue}    \\
\textbf{Block Actions}       & \multicolumn{2}{l}{grasp(block), place(block, on\_block(s))}    \\
\textbf{Location Action}       & \multicolumn{2}{l}{move\_to(home)}    \\
\textbf{Typical Example Timeline}  & \multicolumn{2}{l}{18.6s duration, 186 frames, 10Hz}\\
\cmidrule(r){1-3}
% \midheader{5}{IV-IV Compounds}
\multicolumn{3}{c}{\textbf{3D Coordinate Poses Recorded}}\\
% \midrule
gripper base and center      & rgb camera          & depth camera     \\
robot joints & AR tags + ID\#      & colored blocks       \\
% % \cmidrule(r){1-3}
% % \midheader{5}{IV-IV Compounds}
% \multicolumn{3}{l}{\textbf{3D coordinate poses recorded:}}\\
% % \midrule
% \multicolumn{3}{l}{gripper base and center, camera, joints, AR tags + ID, colored blocks}\\
% \midrule
% \cmidrule(r){1-3}
% %\textbf{Objects present}
% & \textbf{blocks} & \textbf{blocks + toys} \\
% Attempts      & 5,884    & 4,741      \\
% Successes      & 2,451    & 749      \\
% Failures, all kinds      & 3,433    & 5,358      \\
% Failures without errors      & 1233    & 7,088      \\
% Failures with errors    & 2,200    & 1,730      \\
% \multicolumn{3}{l}{\textbf{Success Only Subset}}\\
% Training  &    2,195     & 620       \\
% Validation       & 128     & 64      \\
% Test       & 128     & 64      \\
\bottomrule
\\
\multicolumn{3}{c}{\textbf{Examples by Category out of the Combined Total of 11,977}}\\
\end{tabular}
% \vspace{0.2cm}
\includegraphics[width=1.0\columnwidth]{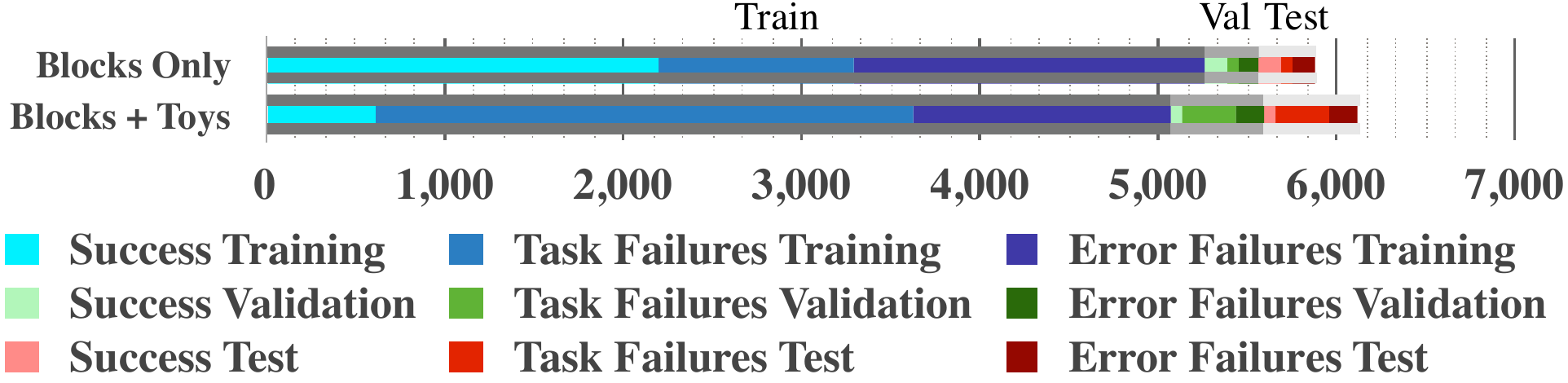}
\caption{
\label{table:dataset} 
% Summary of data available in the CoSTAR block stacking dataset shown in Fig. \ref{fig:workspace}. 
Stacking was conducted under 2 conditions: (1) blocks only and (2) blocks with plush toy distractors. Task Failures complete 5 actions but are unsuccessful at stacking. Causes of failures with errors include security stops from collisions, user emergency stops, planning failures, and software runtime errors.
% Counts for data with any images present
% python view_convert_dataset.py --path ~/.keras/datasets/costar_block_stacking_dataset_v0.4/blocks_only/ --avg_frame_count
% Exact counts for costar dataset version 0.4
% blocks_only Run complete! Counted 5870 files. Total = 1009891 Average = 172
% blocks with plush: Run complete! Counted 6107 files. Total = 1213467 Average = 198
% $ (1009891 + 1213467) / (5870 + 6107)                      
% 185.6356349670201
% Combined total frames:
% 2,223,358
% TODO(ahundt) move plush failures without errors from when gripper would not open to errors category
% block stacking dataset no plush:
% 8541 files
% 2589 with no images
% 22 unreadable
% 5,930 readable examples with data + images
}
\vspace{-0.4cm}
\end{table}

%%%%%%%%%%%%%%%%%%%%%%%%%%%%%%%%\ 
%\section{Related and Prior Work}
%%%%%%%%%%%%%%%%%%%%%%%%%%%%%%%%
\section{Overview and Related Work} \label{ref:prior}

Block stacking is itself already studied to improve scene understanding~\cite{lerer2016learning}, and our videos include stacks standing, leaning, and tumbling. 
This pairs well with ShapeStacks\cite{groth2018shapestacks} a synthetic dataset for understanding how stacks of simple objects stand or fall.
Example use cases for their dataset with our own includes the evaluation of model based methods' ability to accurately predict future consequences and detect subtle collision scenarios with or without an object model.

Intuitively, block stacking might appear to be trivially solved by existing grasping or 3D object pose estimation algorithms alone.
Recent advances in deep learning have revolutionized robotic grasping with perception based methods learned from big data~\cite{levine2016learning, mahler2017dex, levine2015deepvisiomotor,DBLP:journals/corr/PintoG15,levine2017semanticgrasping, agrawal2016learning,redmon2015real}.
One notable limitation of past approaches to robotic manipulation is the restriction of end effector poses to be either vertical and facing down or normal to local depth values, with only an angular parameter available to define orientation changes~\cite{levine2016learning, mahler2017dex, 2018qtopt, 2018grasploop}. 
Also common is the use of depth-only data~\cite{mahler2017dex} which precludes the possibility of object discrimination based on color. 
% Furthermore, such analysis may have been performed in lighting conditions that are very consistent, which can lead to failures in the presence of direct sunlight. 
Progress towards semantic grasping of specific objects \cite{levine2017semanticgrasping} is substantial, but it remains an open problem.

We demonstrate one specific starting condition for the block stacking task, visualized in Fig. \ref{fig:gripper_asymmetry_time_dependencies}, where obstacles and task requirements imply these methods are not sufficient on their own.
We can reach two conclusions from the physical shape, translation, and orientation dependencies in this figure: 
(1) The 4 DOF $(x, y, z, \theta)$ available for gripper motion in current grasping networks \cite{zhang2018real, 2018qtopt, danielczuk2019mechanical} is not sufficient for precisely grasping and then placing one specific block on another in the general case, so at least one additional axis of rotation is necessary.
(2) A neural network which predicts a 6 DOF object pose alone is not sufficient to overcome obstacles because a one to one mapping between 3D object poses and a sequence of successful gripper grasp and placement poses does not exist. 
In this example, no single definition of object poses will work because the required 45\degree rotation for precise placement will not match any square object pose. 
This means that even if an oracle provides both high level task instructions and perfect 3D object poses, an agent must discern the sequence of gripper poses, the shape of the objects, and control the robot correctly without a fatal collision for all time steps in between. 
We describe several of these requirements and challenges in Fig. \ref{fig:StackingSkillsNeuralNetwork}.

These principles extend trivially to 3D if we consider all possible positions and orientations of 4 blocks and all 4 side walls, with rounded wall intersections.
Consideration of other initial states will reveal other clear counter-examples, but we leave this exercise to the reader.
It quickly becomes clear that in the general 3D case of this scenario at least 5 DOF are necessary to successfully grasp and then place a specific block on another. 
Our goal is to eventually generalize to more complex tasks than block stacking, so we design our example algorithm for full 6 DOF gripper motions.

Other work has investigated learning from simulation and then incorporating those models into robotic control \cite{mordatch2016onlinepolicyhumanoid}. 
% Our prior work demonstrates how synthetic datasets with rendered objects mixed with real scene backgrounds combined with deep supervision is an effective approach to generating large training datasets at extremely low cost~\cite{li2017objectparsing}.
Authors have explored simulation \cite{tobin2017simtorealworld,zhang2016reaching, bousmalis2018using} and image composition \cite{li2017objectparsing} to generate training data that transfers to physical scenarios.
Our approach to motion learning is inspired by~\cite{levine2016learning,levine2017semanticgrasping},
with additional extensions based on~\cite{he2016resnetv2,li2017objectparsing,devin2016modular,wang2015training}.
Others have used reinforcement learning for generating API calls for pre-programmed actions~\cite{xu2018neural}.

% The AR tags on the robot are used to perform dual quaternion hand-eye calibration before the dataset was collected, and the AR tag in the bin was used to initialize the table surface for data collection as described in~\cite{paxton2017costar}. Object models and AR tags are not utilized in our example neural network.

% \subsection{Use cases}
An investigation into the multi-step retrieval of an occluded object called  Mechanical Search~\cite{danielczuk2019mechanical} specifically states: ``[The] performance gap [between our method and a human supervisor] suggests a number of open questions, such as: Can better perception algorithms improve performance? Can we formulate different sets of low level policies to increase the diversity of manipulation capability? Can we model Mechanical Search as a long-horizon POMDP and apply deep reinforcement learning in simulation and/or in physical trials to learn more efficient policies?''
Our CoSTAR dataset is specifically designed as a resource for exploring, addressing, pre-training, and benchmarking solutions to questions like these. 

We can also imagine many additional topics for which the JHU CoSTAR Block Stacking Dataset might be utilized, such as the investigation and validation of task and motion planning algorithms offline on real data,
imitation learning and off policy training of Reinforcement Learning\cite{2018qtopt,aytar2018playinghardyoutube} algorithms for complete tasks with a sparse reward, model based algorithms which aim to complete assembly tasks in cartesian or joint space.
Both the dataset and HyperTrees might also be useful for developing, evaluating and comparing algorithms utilizing sim-to-real transfer, GANs, domain adaptation, and metalearning\cite{james2018rcansimtoreal, finn2017model}.
These applications become particularly interesting when the 3D models we have available are added to a simulation, or when this dataset is combined with other real or synthetic robotics datasets.

Finally, Neural Architecture Search is an emerging way to automatically optimize neural network architectures to improve the generalization of an algorithm. Key examples include NASNet~\cite{2017nasnet}, and ENAS~\cite{2018enas}, but a broad overview is outside the scope of this paper, so we refer to a recent survey \cite{2018nassurvey}.

%%%%%%%%%%%%%%%%%%%%%%%%%%%%%%%%
\section{ Block Stacking Dataset }
%%%%%%%%%%%%%%%%%%%%%%%%%%%%%%%%

\label{ssec:datacollection}

\begin{table*}\centering
\ra{1.3}
\setlength\tabcolsep{2pt}
\begin{tabular}{| L | C | C | C | C | C | C | C | C | C | C | C | C | C | C | C | C | C |}
\hline                                           % 1        2        3        4        5        6       7         8         9      10      11     12     13        14      15         16  
\rowcolor{white} Robot Dataset              & \thead{Real\\Data} & \thead{Scene\\Varies} & \thead{Human\\Demo} & \thead{Open\\License} & \thead{Grasp} & \thead{Place} & \thead{Specific\\Objects} & \thead{Scene\\Obstacle} & \thead{Phys.\\Dep.} & \thead{Robot\\Model} & \thead{Val\\Set} & \thead{Test\\Set} & \thead{Code\\Incl.} & \thead{Trials} & \thead{Time\\Steps} & \thead{Rate\\Hz} \\
\Xhline{2\arrayrulewidth}                                          % 1        2          3      4         5         6       7        8        9        10       11       12     13       14     15         16    
\rowcolor{white} \textbf{JHU CoSTAR Block Stacking}         & \cmark & \cmark & \xmark & \cmark & \cmark & \cmark & \cmark & \cmark & \cmark & \cmark & \cmark & \cmark & \cmark & 11,977 & 186 & 10\\
Google Grasping\cite{levine2016learning}   & \cmark & \cmark & \xmark & \cmark & \cmark & \xmark & \xmark & \xmark & \xmark & \xmark & \xmark & \xmark & \xmark & $\sim$800k  & $\sim$25 & 1\\
\rowcolor{white} MIME\cite{sharma2018multiple}              & \cmark & \cmark & \cmark &   --   & \cmark & \cmark & \xmark & \xmark & \cmark & \cmark & \xmark & \xmark & \xmark & 8,260  & $\sim$100  & 7\\
BAIR Pushing\cite{ebert2017self}           & \cmark & \xmark & \xmark &   --   & \xmark & \xmark & \xmark & \xmark & \cmark & \cmark & \xmark & \cmark & \cmark & 45,000  & 30 & --\\
\rowcolor{white} BAIR VisInt-solid/cloth\cite{ebert2018robustness}& \cmark & \xmark & \xmark &   --   & \cmark & \xmark & \xmark & \xmark & \cmark & \cmark & \cmark & \cmark & \cmark & 16k/31k  & 30/20 & --\\
Jacquard\cite{2018jaquardgrasp}            & \xmark & \xmark & \xmark & \xmark & \cmark & \xmark & \xmark & \xmark & \xmark & \cmark & \xmark & \xmark & \xmark & 54,485 & 1 & --\\
\rowcolor{white} Cornell\cite{cornellgrasping}              & \cmark & \xmark & \xmark &   --   & \cmark & \xmark & \cmark & \xmark & \xmark & \cmark & \xmark & \xmark & \cmark & 1,035  & 1 & --\\
Dex-Net 2.0\cite{mahler2017dex}            & \xmark & \xmark & \xmark &   --   & \cmark & \xmark & \xmark & \xmark & \xmark & \cmark & \xmark & \xmark & \cmark & 6.7M  & 1 & --\\
\hline
\end{tabular}

\caption{\label{table:datasetcomparison} A comparison of robotics datasets. Our CoSTAR dataset also includes methods, documentation, examples, and the details to reproduce it. A dash indicates not available or not applicable. Physical dependencies are described in Fig. \ref{fig:gripper_asymmetry_time_dependencies}.
The bin is our ``Scene Obstacle''; forceful collision causes a security stop and the ``Failure with errors'' condition in Table \ref{table:dataset}.
}
\vspace{-0.4cm}
\end{table*}

We define a block stacking task where a robot attempts to stack 3 of 4 colored blocks in a specified order. The robot can be seen in Fig. \ref{fig:workspace}, and examples of key image frames for two stack attempts are  shown in Fig. \ref{fig:attempt}. A dataset summary can be found in Table \ref{table:dataset}.

Data is collected utilizing our prior work on the collaborative manipulation system CoSTAR~\cite{paxton2017costar,paxton2018evaluating}.
CoSTAR is a system designed for end-user creation of robot task plans that offers a range of capabilities plus a rudimentary perception system based on ObjRecRANSAC.
Motion is executed by first planning a direct jacobian pseudoinverse path, with an RRT-connect fallback if that path planning fails.
In a single stack attempt the robot aims to complete a stack by performing 5 actions: 2 repetitions of the CoSTAR \texttt{SmartGrasp} and \texttt{SmartPlace} actions, plus a final move to the home position above the bin. 
The sequence pictured in Fig. \ref{fig:StackingSkillsNeuralNetwork} consists of the following 5 actions from top to bottom: \texttt{grasp(red)}, \texttt{place(red, on\_blue)}, \texttt{grasp(yellow)}, \texttt{place(yellow, on\_red\_blue)}, and \texttt{move(home)}. There are a total of 41 possible object-specific actions: grasp actions interact with each of the 4 colored blocks (4 actions), placement actions are defined for ordered stacks with up to height 2 (36 actions), and \texttt{move(home)}.

The dataset provides the appearance of smooth actions with the gripper entering the frame, creating a stack in the scene, and finally exiting the frame at the end. During real time execution the robot (1) proceeds to a goal, (2) saves the current robot pose, (3) stops recording data, (4) moves out of camera view to the home position, (5) estimates the block poses, (6) moves back to the saved pose, (7) resumes recording, (8) starts the next action. After moving to the final home position object poses are estimated and the maximum $z$ height of a block determines stack success which is confirmed with human labeling. Some features, such as collision checks, are disabled so that a set of near-collision successes and failures may be recorded.

%%%%%%%%%%%%%%%%%%%%%%%%%%%%%%%%
\section{ Problem and Approach }
%%%%%%%%%%%%%%%%%%%%%%%%%%%%%%%%
\label{sec:problemandapproach}

We explore one example application on the CoSTAR dataset by demonstrating how high level pose goals might be set without object models.
We assume that a higher level oracle has identified the next necessary action, and the purpose of the neural network is to learn to set 3D pose goals from data and an object-specific action identifier.
The proposed goal can then be reached by a standard planning or inverse kinematics algorithm. 
The high level task and requirements placed on the network are outlined in Fig. \ref{fig:StackingSkillsNeuralNetwork}.

\begin{figure*}[bt!]
\centering
\includegraphics[width=1\textwidth]{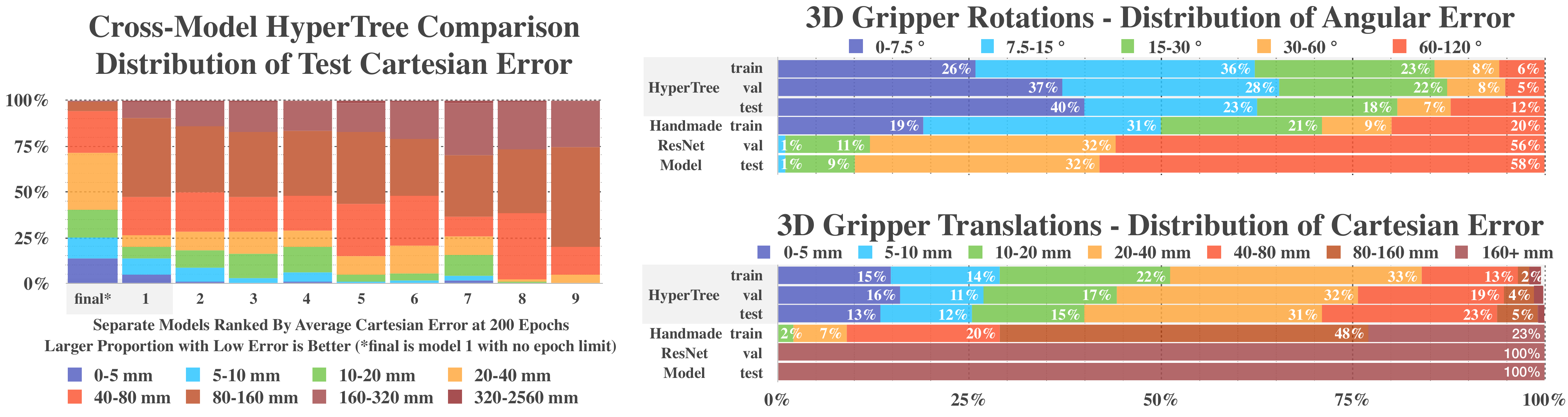}
\caption{ \textbf{(All)} The best models' predictions $P_t$ against ground truth $G_t$ at random times $t$. A high percentage of samples with low error is better. \textbf{(Left)} The importance of hyperparameter choice is visible in models 1-9 which were selected from the best of 1100 HyperTree candidates and then trained for 200 epochs. \textbf{(Top)} Distribution of angular error between predicted and actual 3D gripper rotations $\Delta Rot(r^p_t, r^g_t)$ (Eq. \ref{eq:goal}, and Fig. \ref{fig:workspace}). \textbf{(Bottom)} Distribution of translation error $\|v^p_t-v^g_t\|$ (Eq. \ref{eq:goal}, and Fig. \ref{fig:workspace}). 
}
\label{fig:accuracy3d}
\vspace{-0.6cm}
\end{figure*}

\subsection{Goals and Encodings}
\label{ssec:goalsandencodings}
Each successful stacking attempt consists of 5 sequential actions (Fig. \ref{fig:StackingSkillsNeuralNetwork}, \ref{fig:attempt}) out of the 41 possible object-specific actions described in Sec. \ref{ssec:datacollection}. 
% Predictions $P_t$ are made with respect to ground truth goal poses $G_t$ at the time $g$ end of each action. 
% Each $G_t$ is extracted based on the frame $g$ at which at which a gripper open or close motion begins. 
Stacking attempts and individual actions vary in duration and both are divided into separate 100 ms time steps $t$ out of a total $T$. 
There is also a pose consisting of translation $v$ and rotation $r$ at each time step (Fig. \ref{fig:workspace}), which are encoded between [0,1] for input into the neural network as follows: 
\label{eq:translation}
The \textbf{translation vector encoding} is $
v = (x, y, z)/d+0.5$, where $d$ is the maximum workspace diameter in meters. The \textbf{Rotation $r$ axis-angle encoding} is 
\label{eq:rotation}
$r = (a_x, a_y, a_z, sin(\theta), cos(\theta))/s + 0.5$, where $a_x,a_y,a_z$ is the axis vector for gripper rotation, $\theta$ is the angle to rotate gripper in radians, and $s$ is a weighting factor relative to translation.
\noindent
\textbf{Example $E$ is the input to the neural network}: 
\begin{equation}
E_t = (I_0,I_t, v_t, r_t, a_t) 
\label{eq:example}
\end{equation}
Where $I_0$ and $I_t$ are the initial and current images, $v_t$, $r_t$ are the respective base to gripper translation and rotation (Fig. \ref{fig:workspace}). $a_t$ is the object-specific one-hot encoding of 41 actions. \textbf{Ground Truth Goal Pose $G_t$} from Fig. \ref{fig:workspace} is the 3D pose at time $g$ at which the gripper trigger to open or close, ending an action in a successful stacking attempt:
\begin{equation}
G_t = (v^g_t, r^g_t) | t \leq g \leq T, e_g \neq e_{g-1}, a_g == a_t
\label{eq:goal}
\end{equation} 
where $g$ is the first time the gripper moves after $t$, $e$ is the gripper open/closed position in [0, 1]. 
% The goal is defined by when the gripper position changes. 
Finally, the
\textbf{Predicted Goal Pose $P_t = (v^p_t, r^p_t)$} is a prediction of $G_t$.

Each example $E_t$ has a separate sub-goal $G_t$ defined by (1) the current action $a_t$ and (2) the robot's 3D gripper pose relative to the robot base frame at the time step $g$ when the gripper begins moving to either grasp or release an object. Motion of the gripper also signals the end of the current action, excluding the final \texttt{move(home)} action, which has a fixed goal pose.

% \footnote{For videos of CoSTAR and the stacking dataset see: \\ \url{https://www.youtube.com/playlist?list=PL7hJDTQ2ZjSvsSJk7oTVti7udcBEcIHof}

% % , \url{https://youtu.be/PgzF3yaxNhE}  
% % and \url{https://www.youtube.com/playlist?list=PLF86ez-NVmyFMuj10dkUkgGlGpcM5Vok9}
% .}

%%%%%%%%%%%%%%%%%%%%%%%%%%%%%%%%%%%%%%%%%%%%%%%%%%%%%%%%%%%%%%%%%
\subsection{Exploring the Block Stacking Dataset}
\label{exploringdataset}

\begin{figure}[bt!]
    \centering
    % \hfill
    \includegraphics[width=\columnwidth]{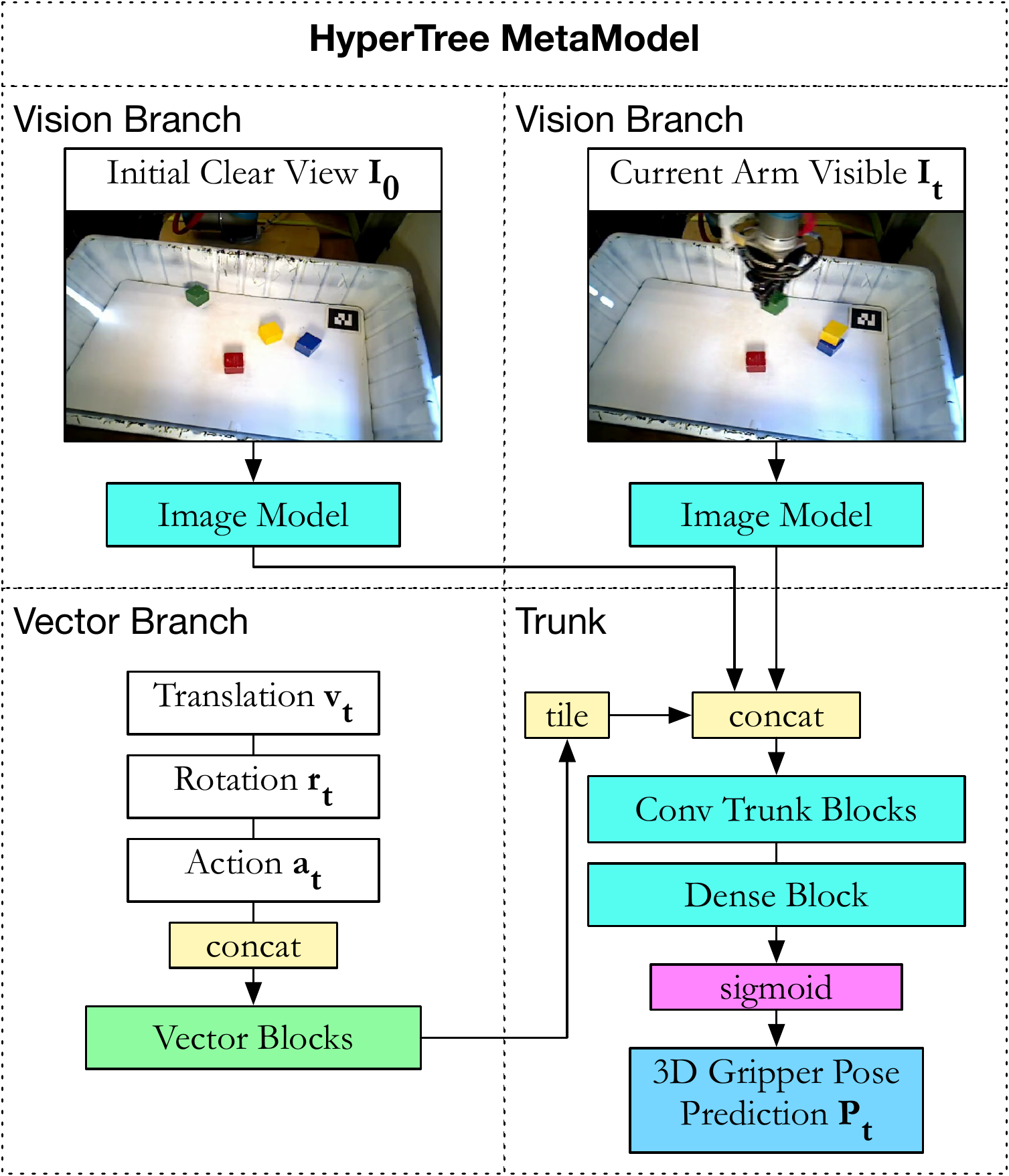}
    \caption{
    \label{fig:MotionNetwork} %Task Action Proposal Network:
    A detailed view of the HyperTree MetaModel configured for predicting 3D ground truth goal poses, $G_t$, on the block stacking dataset. HyperTrees can accept an arbitrary number of image and vector inputs. Hyperparameter definitions are in Table \ref{table:HyperParameters}. ``Blocks'' are a sequence of layers.
    % Details for the dataset are in Fig. \ref{fig:workspace}, \ref{fig:attempt}, and Table \ref{table:dataset}.
}
\vspace{-0.4cm}
\end{figure}

% To establish an initial baseline we first surveyed existing work available at the time. 
% Finally, in the case of vector outputs there may be a pooling operation followed by activation, or the latent variables may simply be activated at each location in the case of pixel-wise outputs.
We implemented several models similar to those found in existing work\cite{levine2016learning, Kumra_2017_graspcnn, he2016resnetv2, 2016densenet}. 
We minimized our modifications to those necessary to accommodate our data encoding.
% However, these exact models cannot be used directly for this task because inputs to the network include RGB images, the current gripper pose, and the action to take.
Despite our best efforts, no baseline model we tried, and no hand-made neural network variation thereof could converge to reasonable values.
% For this reason, we investigated various variations thereof which incorporate ResNetv2\cite{he2016resnetv2}, and DenseNet\cite{2016densenet} with which we also made no significant progress.
Once we verified the CoSTAR dataset was itself correct, evaluated models on the Cornell Grasping Dataset\cite{cornellgrasping} without issue, and tried a variety of learning rates, optimizers, models and various other parameters tuned by hand this complete lack of progress became very surprising.
We analyze the underlying cause in Sec. \ref{ssec:hypertreeablation} and include one reference model based on Kumra et. al.\cite{Kumra_2017_graspcnn} in Fig. \ref{fig:accuracy3d} for comparison.
It quickly became clear that manually tweaking configurations would not be sufficient, so a more principled approach to network design would be essential.
To this end, Neural Architecture Search and hyperparmeter search are well studied methods for automatically finding optimal parameters for a given problem, and we apply them here.

\subsection{HyperTree MetaModel}
\label{ssec:hypertreeMetaModel}
\label{ssec:hypertreesearch}

Much like how Dr. Frankenstein's creature was assembled from pieces before he came to life in the eponymous book, HyperTrees combine parts of other architectures to optimize for a new problem domain.
% Robotics based networks often share several common architecture elements.
Broadly, robotics networks often have inputs for images and/or vectors which are each processed by some number of neural network layers. 
These components may then be concatenated to apply additional blocks of layers for data fusion. 
The output of these layers are subsequently split to one or more block sequences, typically dense layers.
% Therefore, to find a neural network suitable to this task we designed the HyperTree Architecture Search Space and MetaModel. 
% We compare HyperTrees to rENAS, our extension of Efficient Neural Architecture Search (ENAS)~\cite{2018enas}.
% We demonstrate HyperTrees on this dataset by predicting full 3D poses semantically for the purpose of grasping and placing specific objects. 
% Architectures fitting this layout have been partially or fully designed by hand in a completely reasonable manner, but is each the optimal choice for that domain?
To search for viable architectures, the HyperTree MetaModel (Fig. \ref{fig:MotionNetwork}) parameterizes these elements (Table \ref{table:HyperParameters}) so that models and their constituent parts might be defined, swapped, evaluated and optimized in a fully automatic fashion. 
In fact, a HyperTree MetaModel's search space can generalize many of the previously referenced architectures as a special case.

\begin{figure}[t]
\centering
\includegraphics[width=1\columnwidth]{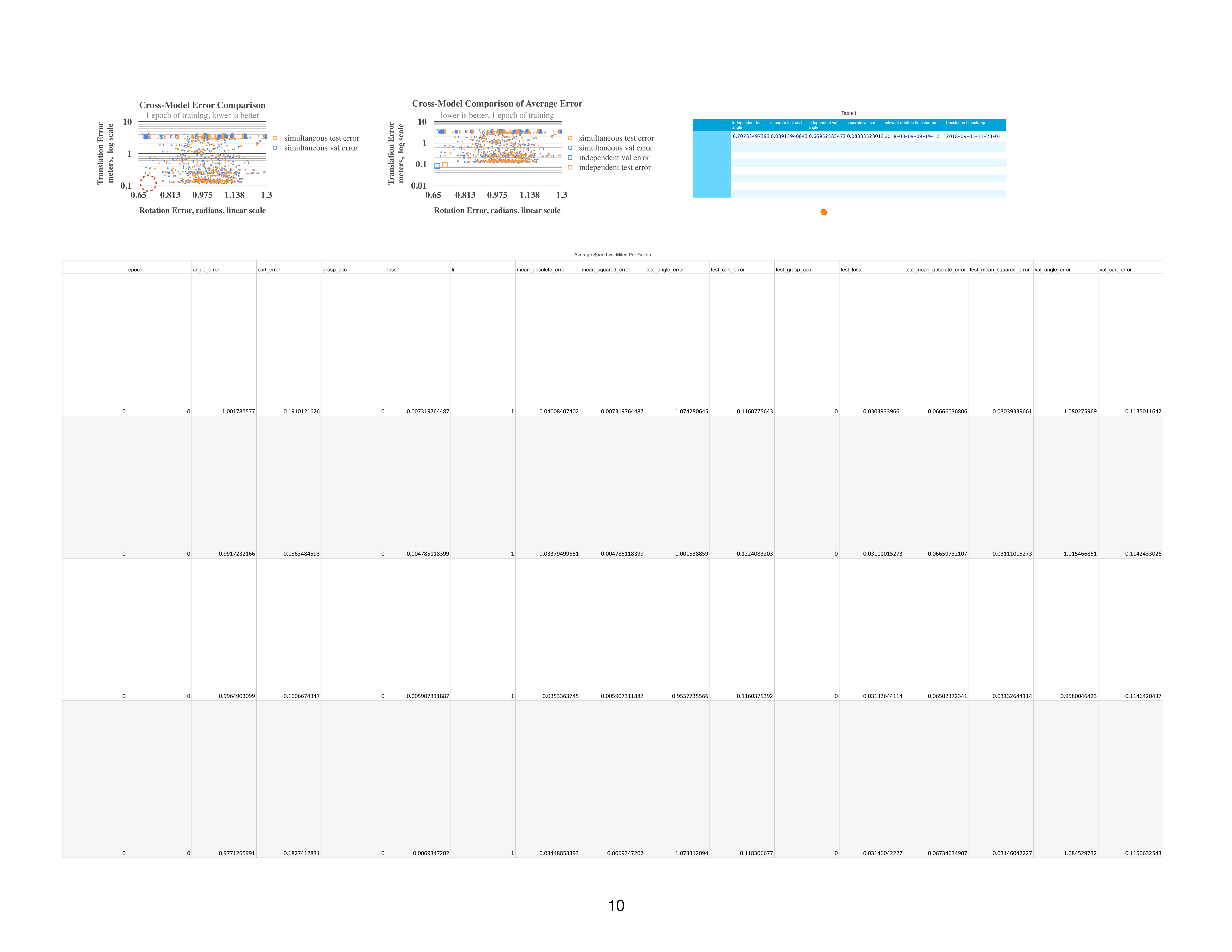}
% source: https://docs.google.com/spreadsheets/d/1nQJvWKrcP1OowQGdnoRqsE6HoDsOoEpMqvL-Q26ehO8/edit?usp=sharing
% \includegraphics[width=\columnwidth]{unrealcv_tools_segmentation}
\caption{ A cross-model comparison of average error with 1 epoch of training. Each dot represents a single HyperTree architecture which predicts both translation and orientation, $P_t$. 
Many models within the search space do not converge to useful predictions. 
The squares demonstrate how a selected pair of HyperTree architectures reduce error by predicting translation $v^p_t$ and rotation $r^p_t$ independently. 
% \cprm{Each one of these models outputs a prediction for $G_t$, including both position and orientation.} 
% A model which accurately predicted both translation and orientation would fall in the empty red circle, and the lack of such a model motivates our decision to define one model which predicts position $v^p_t$ and another which predicts orientation $r^p_t$.
% note: Andrew has verified this chart was generated after the relevant critical bugs in the git history were fixed
}
\label{fig:HyperTreeModelSearchComparisonOfError}
\vspace{-0.2cm}
\end{figure}

\begin{table*}\centering
\ra{1.3}
% \begin{tabular}{@{}lrrrcrrrcrrr@{}}\toprule
\begin{tabular}{LRRRCRRRCRRR@{}}\toprule
% \multicolumn{1}{l}{Hyperparameter} & \multicolumn{1}{c}{Search Space} & \phantom{abc}& \multicolumn{1}{c}{Translation Model} &
% \phantom{abc} & \multicolumn{1}{c}{Rotation Model}\\% \cmidrfule{2-2} \cmidrule{4-4} \cmidrule{6-6}
\rowcolor{white} Hyperparameter & \thead{Search Space} & \phantom{abc}& \thead{Translation Model} &
\phantom{abc} & \thead{Rotation Model}\\% \cmidrfule{2-2} \cmidrule{4-4} \cmidrule{6-6}
%& $t=0$ & $t=1$ & $t=2$ && $t=0$ & $t=1$ & $t=2$ && $t=0$ & $t=1$ & $t=2$\\ 
\midrule
\rowcolor{white} Image Model & [VGG, DN, RN, IRNv2, NAS] && NAS && VGG16\\
Trainable Image Model Weights* & [True, False] && True && True\\
\rowcolor{white} CoordConv Layer Location & [None, Pre-Trunk, Pre-Image] && None && Pre-Trunk\\
Loss Function* & [mse, mae, msle] && mse && msle\\
\rowcolor{white} Activation (Conv3x3, Vector Block, Dense Block) & [relu, elu, linear]&& relu, relu, relu && N/A, relu, relu\\
Vector Block Model & [Dense, DN] && Dense && DN\\
\rowcolor{white} Vector Block Layer Count & $ n \in [0..5)$&& 2 && 1\\
Conv Trunk Block Model & [Conv3x3, NAS, DN, RN] && Conv3x3 && NAS\\
\rowcolor{white} Conv Trunk Block Count & $n \in [0..11)$&& 8 && 8\\
% Trunk Filters & ${2}^{n} | n \in [6..12)$&& TODO&& TODO\\
Filters (Vector, Trunk, Dense Block) & ${2}^{n} | n \in [6..13)$, $[6..12)$, $[6..14)$&& 2048, 1024, 512&& 256, 32, 2048\\
\rowcolor{white} Dense Block Layer Count & $ n \in [0..5)$&& 2 && 3\\
% Top Block Filters & ${2}^{n} | n \in [6..14)$&& TODO&& TODO\\
Normalization (Vector, Trunk) & [Batch, Group, None] && Batch, None && Batch, Batch \\
\rowcolor{white} Optimizer* & [SGD, Adam] && SGD && SGD\\
Initial Learning Rate* & $0.9^n | n \in [0.0 .. 100.0]$ continuous && 1.0 && 1.0\\
\rowcolor{white} Dropout rate* & $[0, 1/8, 1/5, 1/4, 1/2, 3/4]$ && $1/5$ && $1/5$\\
% Preprocessing Mode & [tensorflow, caffe, torch] && tensorflow && tensorflow \\ 
\bottomrule
\end{tabular}
\caption{\label{table:HyperParameters} Architecture Search Parameters for the HyperTree MetaModel defined in Figure \ref{fig:MotionNetwork}. Image Models: VGG16~\cite{simonyan2015very}, DN is DenseNet 121~\cite{2016densenet}, RN is ResNet 50~\cite{he2016resnetv2,he2015resnet}, IRNv2 is Inception ResNetv2~\cite{2016inceptionresnetv2}, NAS is NASNet Mobile~\cite{2017nasnet}. For Conv Trunk Block Model, NAS refers to the NASNet A Cell, DN refers to the DenseNet Dense Block, and ResNet refers to their Identity Block. The Activation hyperparameter applies to the Vector Model, the Conv3x3 Trunk Block, and the Dense Layers in the Dense Block. CoordConv~\cite{2018coordconv} ``Pre-Image'' applies an initial CoordConv Layer to each input image and CoordConv ``Pre-Trunk'' applies a CoordConv layer after the vision and vector branches have been concatenated in the HyperTree Trunk. In Vector Block Model, ``Dense'' is a sequence of Dense Layers, while ``DNBlock'' is a DenseNet style block where Dense layers replace convolutions for the purpose of working with 1D input. Starred * parameters were searched then locked in manually for subsequent searches to ensure consistency across models.}
\vspace{-0.4cm}
\end{table*}

% We define a HyperTree network using Keras~\cite{chollet2015keras} backed by Tensorflow~\cite{tensorflow2015-whitepaper} in search of a model which can most accurately predict the solution to a specific problem. 
We explore and then optimize the models' hyperparameter based configuration of the network structure using the standard optimization framework GPyOpt~\cite{gpyopt2016}. 
We (1) run HyperTree search for 1 epoch on between 500-5,000 models with augmentation, such as cutout\cite{2017cutout}, disabled depending on the available computing resources and dataset size. 
From this we (2) automatically construct a table of the best models, which we sort by a chosen metric, typically the average cartesian or angular validation error. 
% If no models stand out with respect to the validation metrics, we instead sort based on the equivalent training metrics. 
We then (3) conduct a second automated training run proceeding down the top 1-10\% of this sorted list for 10 epochs per model, which is added to our model table. In step (4) we repeat steps 2 and 3 for 200 epochs with 2-10 models and augmentation enabled, if appropriate. Step (5) is a 600 epoch training run initialized with the best model from step 4 resumed as needed until convergence, to reach a final model according to the chosen validation metric. An optional step (6) is to manually narrow the hyperparameter search space to ranges defined by the best image and trunk models and repeat steps 1-5.

Variables, dimensions and inputs above (as in Sec. \ref{ssec:goalsandencodings} and \ref{ssec:hypertreeMetaModel}) are parameterized. For example, HyperTrees accept zero or more vector and image inputs. The Cornell Grasping Dataset provides one image, and we utilize two on the block stacking dataset.
% Include in arxiv version:
% A sample training curve from step 4 with our best translation model is provided in Fig. \ref{fig:trainingplot}. 
Block stacking results are described in Fig. \ref{fig:accuracy3d}, Table \ref{table:HyperParameters}, and Section \ref{sec:results}.

\section{Results}

\label{sec:results}
\textbf{Cornell Grasping Dataset:} We first demonstrate that the HyperTree MetaModel with vector inputs generalizes reasonably well on the Cornell Grasping Dataset. 
Our pose classification model gets 96\% object-wise 5-fold cross evaluation accuracy, compared with 93\% for DexNet 2.0~\cite{mahler2017dex}. 
State of the art is an image-only model at 98\%\cite{zhang2018real}.

\textbf{Separation of translation and rotation models:} In our initial search of the CoSTAR Block Stacking Dataset, a single model contained a final dense layer which output 8 sigmoid values encoding $P_t$. The results of this search represent 1,229 models which are pictured as dots in Fig. \ref{fig:HyperTreeModelSearchComparisonOfError}. 
The figure demonstrates that we found no models which were effective for both translation $v^p_t$ and rotation $r^p_t$ simultaneously. 
This observation led us to conduct independent model searches with one producing 3 sigmoid values $v^p_t$ (Eq. \ref{eq:goal}) encoding translations, and 5 sigmoid values predicting $r^p_t$ (Eq.\ref{eq:goal}) encoding rotations in $P_t$ (Eq. \ref{eq:goal}). 
An example of the resulting improvement in performance plotted as squares is shown in Fig. \ref{fig:HyperTreeModelSearchComparisonOfError}. 

\textbf{CoSTAR Block Stacking Dataset:} The hyperparameters of the best models resulting from the separate translation and rotation model searches are in Table \ref{table:HyperParameters}, while the performance of the top translation and rotation model is detailed in Fig. \ref{fig:accuracy3d} for the training, validation, and test data. Results are presented on the success-only non-plush subset because the plush subset was being prepared during these experiments. For translations on the HyperTree network, 67\% of test pose predictions are within 4 cm and the average error is 3.3 cm. 
% rENAS translations did not perform as well with 24\% of test predictions within 4 cm and an average error of 12 cm. 
For comparison, the colored blocks are 5.1 cm on a side.
% 77\%  of rENAS rotation test prediction are within 15\degree of error, and the average rotation error is 12.6\degree. 
HyperTrees have 81\% of rotation predictions within 30\degree and an average test angular error of 18.3\degree.

\subsection{Ablation Study}
%%%%%%%%%%%%%%%%%%%%%%%%%%%
\label{ssec:hypertreeablation}
In essence, HyperTree search is itself an automated ablative study on the usefulness of each component in its own structure. 
This is because a hyperparameter value of 0 or None in Table \ref{table:HyperParameters} represents the case where that component is removed.
% Specifically, a particular component is skipped when the number of layers in  corresponding to that component has a value of 0 or None. 
For this reason, the best HyperTree models will or will not have these components depending solely on the ranking of validation performance (Fig. \ref{fig:HyperTreeModelSearchComparisonOfError}).
% before final performance is confirmed on the test data. 
For example, a ResNetv2\cite{Kumra_2017_graspcnn} based grasping model like the manually defined one in Fig. \ref{fig:accuracy3d} is a special case which would rise to the top of the ranking if it were particularly effective.
% The one exception is that an Image Model is always defined, but Image Models are already proven to be successful at discriminating objects in a wide variety of domains as described in the Introduction and Related Work sections.

As we look back to our initial hand-designed models (Sec. \ref{exploringdataset}), recall that these did not converge to useful levels of error. 
Fig. \ref{fig:HyperTreeModelSearchComparisonOfError} reveals why this might be. 
Only a select few of the HyperTree models make substantial progress even after training for 1 full epoch of more than 1 million time steps.
Essentially, this means the hand-designed models are simply not converging due to the choice of hyperparameters. 
For this reason we can conclude that an automated search of a well designed search space can improve outcomes dramatically.

An additional HyperTree search of 1100 cartesian models confirms that differences in model quality persist with additional training (Fig. \ref{fig:accuracy3d}). 
This search specified a NASNet-Mobile~\cite{2017nasnet} image model and either a Conv3x3 or NASNet model A cell trunk, selected to explore the space around our final cartesian model. We conducted an initial 1 epoch run, a second 40 epoch run, and then a final 200 epoch run on the 9 best models with respect to validation cartesian error. 
The hyperparameters of the top 9 models vary widely within the search space. Examples of variation include: 0-3 vector branch layers, both vector block models, 0-4 dense block layers, 2-10 trunk layers, 512-8192 vector filters, all 3 CoordConv options, and both trunk options. This dramatic variation is very counter-intuitive.
% The broad range of prediction accuracy across models shown in Fig. \ref{fig:accuracy3d} and \ref{fig:HyperTreeModelSearchComparisonOfError} shows how trivial hand selection of models on a new problem might lead one to conclude neural networks are ineffective, when results actually reflects poor hyperparameter choices. 
Indeed, we found the selection of 8 separate 32 filter NASNet A cells in our own best rotation HyperTree model (Table \ref{table:HyperParameters}) to be a very surprising choice. 
We would be unlikely to select this by hand.
This unpredictability implies that there are many local minima among different possible architectures.
Therefore, the broader conclusion we draw here is that researchers applying neural networks to new methods should perform broad hyperparameter sweeps and disclose their search method before reaching a firm conclusion regarding the strength of one method over another.

% It features a large number of vector filters and a vgg image model which feeds into 8 consecutive nasnet a cells with a mere 32 filters. This transition from large to small filter counts differs substantially from models found in other literature\cite{levine2016learning,2018qtopt,2016inceptionresnetv2} including all the image models in our search space, although this may essentially be analogous to a ResNet\cite{he2015resnet} bottleneck layer.

% Evaluation was performed on 128 randomly selected validation and test examples. Our evaluation results are summarized in Fig. \ref{fig:accuracy3d}. We set a range of thresholds on which we could cumulatively assess predicted positional and rotational accuracy in comparison to our recorded data. At each time step in a successful grasp attempt we found the corresponding robot pose at the goal time step, and compared our error in terms of the cartesian distance and absolute angular distance between our predicted poses and the actual poses of the robot.

% \subsection{Benefits, Limitations and Future Work}
\subsection{Physical Implications and Future Work}

\begin{figure}[t]
    \centering
    \hfill
    \includegraphics[width=\columnwidth]{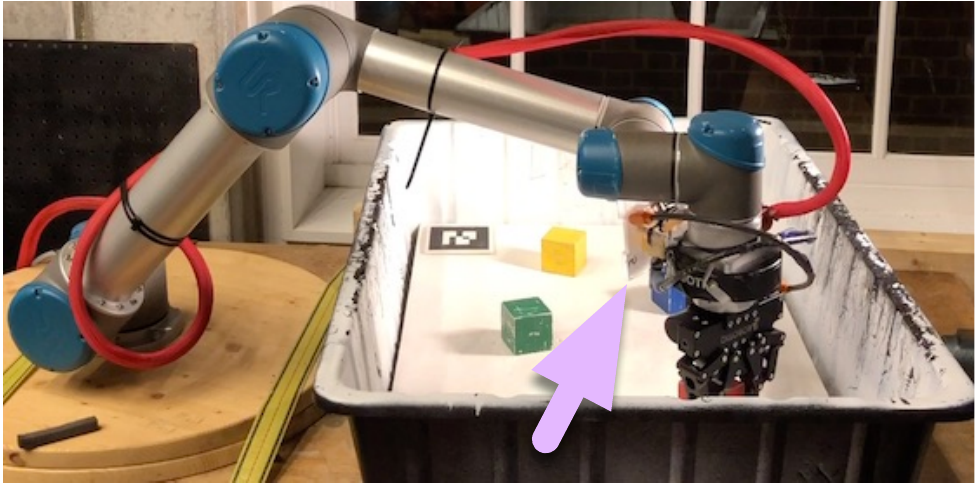}
    \caption{
    \label{fig:example_rotation_ar_tag_facing_away} %Task Action Proposal Network:
    A successful execution of the \texttt{grasp(red)} action with our final model. The predicted gripper orientation keeps the attached AR tag facing away from the walls.
    }%
\vspace{-0.4cm}
\end{figure}

An example of the physical behavior of our final model (Fig. \ref{fig:example_rotation_ar_tag_facing_away}) shows initial progress towards an understanding of the obstacles in the scene, because the protruding side of the gripper faces away from the wall.
Our accompanying video shows several qualitative test grasps and the motion of the predicted pose as a block moves around the scene.
However, these qualitative tests also indicate the current model is not yet accurate enough for end-to-end execution, which we leave to future work.

Several clear avenues for improvement exist.
Predictions might be made on a pixel-wise basis\cite{2018grasploop} to improve spatial accuracy, and pose binning\cite{mahendran2018monocularposebindelta} might improve accuracy.
The Cross Entropy Method could sample around proposals for assessment with a Q function\cite{2018qtopt}.
In turn, a well defined MetaModel based on HyperTrees might improve the accuracy of the networks underlying these other methods.

% Future work remains to fully understand how to benchmark accuracy of grasping tasks when leveraging the combination of simulated and real data. 
Beyond models, several open questions remain before we can more fully leverage datasets:
How can we assess accuracy with respect to successful or failed end-to-end trials without a physical robot?
For example, there is not a trivial mapping from a given rotation and translation error to a trial's success, so what metric will best generalize to real robot trials?
Can we encode, embed, represent, and evaluate such information in a way that generalizes to new situations?
% Should a neural network based classifier or a model based algorithm evaluate proposals which differ from the ground truth, or is there some third approach that ties simulation in with the real physical data?
The CoSTAR dataset can itself serve as a medium with which to tackle these objectives.

%%%%%%%%%%%%%%%%%%%%
\section{Conclusion}
%%%%%%%%%%%%%%%%%%%%

We have presented the CoSTAR Block Stacking Dataset as a resource for researchers to investigate methods for perception-based manipulation tasks. This dataset supports a broad range of investigations including training off-policy models, the benchmarking of model based algorithms against data driven algorithms, scene understanding, semantic grasping, semantic placement of objects, sim-to-real transfer, GANs, and more.
% The CoSTAR Block Stacking Dataset
The CoSTAR BSD can serve to bridge the gap between basic skills and multi-step tasks, so we might explore the broader capabilities necessary to achieve generalized robotic object manipulation in complex environments.

To establish a baseline for this dataset we created the HyperTree MetaModel automated search method, which is designed for this problem and others in which existing architectures fail to generalize.
Our final model from this search qualitatively demonstrates grasping of a specific object and can correctly avoid a scene's boundaries, an essential capability for the full stacking task in a real-world environment.

\section {Acknowledgements}

We thank Chunting Jiao for his assistance with data collection. This material is based upon work supported by the National Science Foundation under NSF NRI Grant Award No. 1637949.

\bibliography{sample}
%\printbibliography

\appendix

\subsection{Goals and Encoding Details, expanded}

Each successful stacking attempt consists of 5 sequential actions (Fig. \ref{fig:StackingSkillsNeuralNetwork}, \ref{fig:attempt}) out of the 41 possible object-specific actions described in Sec. \ref{ssec:datacollection}. 
Stacking attempts and individual actions vary in duration and both are divided into separate 100 ms time steps $t$ out of a total $T$. 
There is also a pose consisting of translation $v$ and rotation $r$ at each time step (Fig. \ref{fig:workspace}), which are encoded between [0,1] for use in a neural network as follows:
%%%%%%%%%%%%%%%%%%%%%%%%%%%%%%%%%%%%%%%%%%

\noindent
\textbf{Translation $v$ vector encoding:}
\begin{equation}
v = (x, y, z)/d+0.5
\label{eq:translation_appendix}
\end{equation} 
% http://tex.stackexchange.com/a/95842/41590
% note: no empty newlines here, that will change indentation
\begin{conditions}
 d               & 4, max workspace diameter (meters) \\
 x,y,z           & robot base to gripper tip translation (meters) \\
 v               & array of 3 float values with range [0,1] \\
\end{conditions}
%%%%%%%%%%%%%%%%%%%%%%%%%%%%%%%%%%%%%%%%%%
\noindent
\textbf{Rotation $r$ axis-angle encoding:}
\begin{equation}
r  = (a_x, a_y, a_z, sin(\theta), cos(\theta))/s + 0.5
\label{eq:rotation_appendix}
\end{equation} 
% http://tex.stackexchange.com/a/95842/41590
% note: no empty newlines here, that will change indentation
\begin{conditions}
 a_x,a_y,a_z               & axis vector for gripper rotation \\
 \theta               & angle to rotate gripper (radians) \\
 s               & 1, scaling factor vs translation \\
 r               & array of 5 float values with range [0,1]
\end{conditions}
%%%%%%%%%%%%%%%%%%%%%%%%%%%%%%%%%%%%%%%%%%
%%%%%%%%%%%%%%%%%%%%%%%%%%%%%%%%%%%%%%%%%%
\noindent

%%%%%%%%%%%%%%%%%%%%%%%%%%%%%%%%%%%%%%%%%%

\noindent
\textbf{Example $E$ is the input to the neural network} defined at a single time step $t$ in a stacking attempt:
\begin{equation}
E_t = (I_0,I_t, v_t, r_t, a_t) 
\label{eq:example_appendix}
\end{equation} 
% http://tex.stackexchange.com/a/95842/41590
% note: no empty newlines here, that will change indentation
\begin{conditions}
 T               & Total time steps in one stack attempt \\
 t               & A single 100ms time step index in $T$ \\
 v_t             & Base to gripper translation, see Eq. \ref{eq:translation_appendix} \\
 r_t             & Base to gripper rotation, see Eq. \ref{eq:rotation_appendix} \\
 h, w, c         & Image height 224, width 224, channels 3\\
 \matrixdim{I}{h \times w \times c} & RGB image tensor scaled from -1 to 1 \\
 I_0             & First image, clear view of scene, $t=0$  \\
 I_t             & Current image, robot typically visible \\
 K, k            & 41 possible actions, 1 action's index \\
 \matrixdim{a_t}{1 \times K} & action one-hot encoding \\
%  a_t[k]          & 1 if an action is active, 0 otherwise \\
\end{conditions}

\noindent
\textbf{Ground Truth Goal Pose $G_t$} from Fig. \ref{fig:workspace} is the 3D pose time $g$ at which the gripper trigger to open or close, ending an action in a successful stacking attempt:
\begin{equation}
G_t = (v_g, r_g) | t \leq g \leq T, e_g \neq e_{g-1}, a_g == a_t
\label{eq:goal_appendix}
\end{equation} 
% http://tex.stackexchange.com/a/95842/41590
% note: no empty newlines here, that will change indentation
\begin{conditions}
  g  & First time the gripper moves after $t$ \\
  e  & gripper open/closed position in [0, 1] \\
  e_g \neq e_{g-1} & gripper position changed \\
  a_g == a_t & action at time $t$ matches action at time $g$ \\
  G_t & array of 8 float values with range [0,1] \\
  (v^g_t, r^g_t) & goal pose, same as $(v_g, r_g)$ \\
%   (v^p_t, r^p_t) & $P_t$ is a prediction of $G_t$ \\
\end{conditions}

\noindent
\textbf{Predicted Goal Pose $P_t = (v^p_t, r^p_t)$} is a prediction of $G_t$.

Each example $E_t$ has a separate sub-goal $G_t = (v^g_t, r^g_t)$ defined by (1) the current action $a_t$ and (2) the robot's 3D gripper pose relative to the robot base frame at the time step $t$ when the gripper begins moving to either grasp or release an object. Motion of the gripper also signals the end of the current action, excluding the final \texttt{move(home)} action, which has a fixed goal pose.
% The fixed final goal pose creates a complete task loop, thus it may be possible to simulate retries with the dataset.

\subsection{CoSTAR Block Stacking Dataset Details}

% If a sequence of actions is completed successfully without system errors and the traditional CoSTAR perception attempts to determines the height of the highest block, which indicates that a stack has been completed successfully. If successful, a file is saved indicating \texttt{success}, otherwise it is saved indicating \texttt{failure}. If any system errors occur, such as a security stop or motion planning failure, the example is saved as an \texttt{error.failure}.

We will outline a few additional CoSTAR BSD details here, and you can find our full documentation and links to both tensorflow and pytorch loading code at \url{sites.google.com/site/costardataset}. We include extensive notes with the dataset, explaining specific events and details of interest to researchers but outside the scope of this paper, such as where to obtain simulated robot models and dates when part failures occurred.
We have included certain details regarding the approach, data channels, update frequency, time synchronization, and problems encountered throughout the data collection process in Fig. \ref{table:dataset} that are not part of our approach to goal pose prediction, but may be useful for an approach to the stacking problem that is completely different from our own.
We have also tried to ensure sufficient data is available for tackling other perception and vision related challenges.
Attention to these details ensure a robotics dataset might prove useful as a benchmark for future research with methods that differ substantially from the original paper. 

In between stack attempts the robot returns to its past saved poses in an attempt to unstack the blocks to automatically reset itself for the next attempt. If too many sequential errors are encountered the data collection application restarts itself which mitigates most system state and traditional motion planning errors. With this approach we find that we can automate the collection of stack attempts with approximately 1 human intervention per hour due to incidents such as security stops and failure cases in which all objects remain exactly where they started. A successful stack attempt typically takes on average about 2 minutes to collect and contains about 18 seconds of data logged at 10 Hz (100ms time steps), but this figure varies substantially across examples.

The AR tags on the robot is used to perform dual quaternion hand-eye calibration before the dataset was collected, and the AR tag in the bin was used to initialize the table surface for data collection as described in~\cite{paxton2017costar}. 
Object models and AR tags are not utilized in the neural network.

\begin{figure*}[btp!]
\centering
\includegraphics[width=0.8\textwidth]{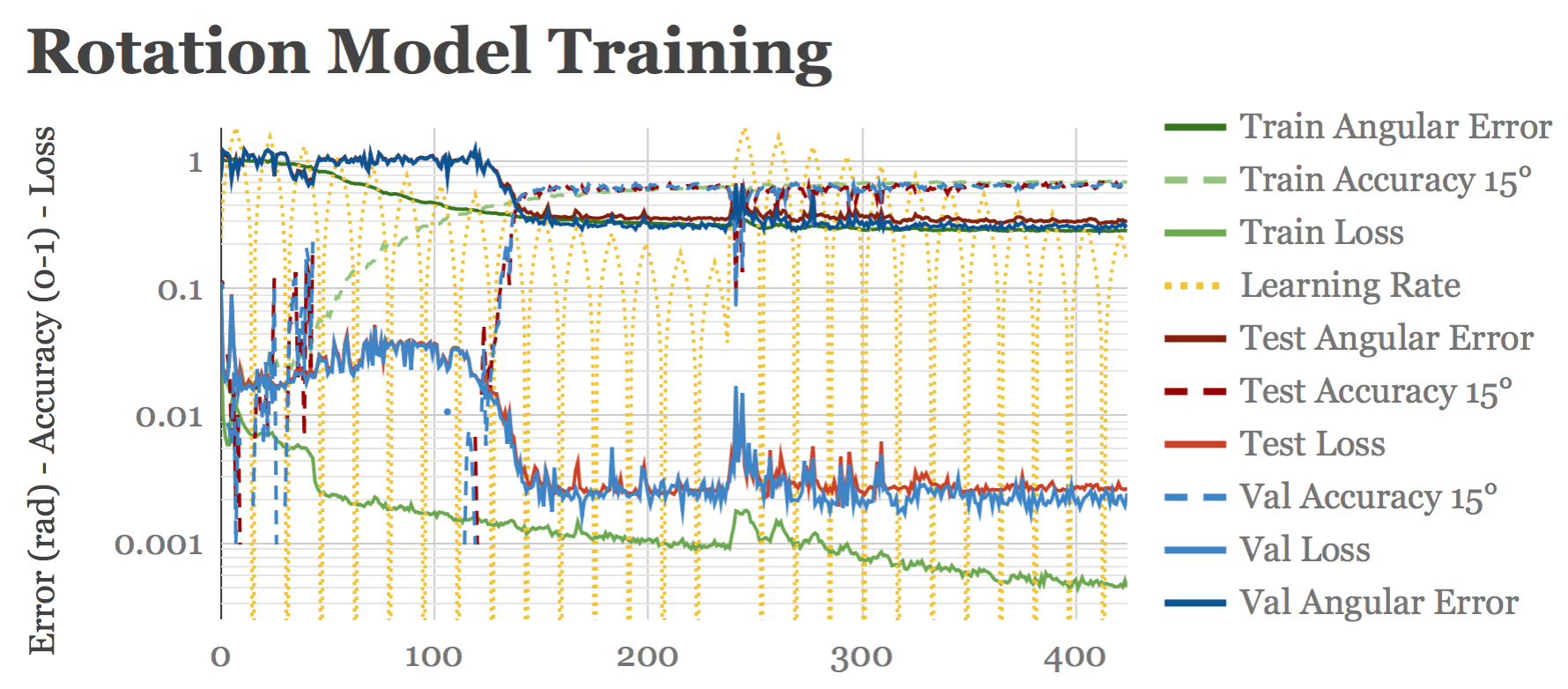}
% https://docs.google.com/spreadsheets/d/e/2PACX-1vRRPOWTGuMHP-KKTvvMP3-3TmBXdoTGnJFfKvdIDaSS0ZJT-zM0qejGwES5-mi5MmCN3pdr43hTsfKk/pubchart?oid=154282479&amp;format=interactive
\caption{\label{fig:trainingplot} Final training of the HyperTree rotation model in Table \ref{table:HyperParameters} and Fig. \ref{fig:accuracy3d}. Higher accuracy, lower error,  and lower loss is better. Training was restarted on epoch 238, with a corresponding increase in learning rate. The final result is average angular errors of $16.0\degree$(val) and $18.3\degree$(test) at epoch 411. The horizontal axis represents performance at each training epoch on a linear scale while the vertical axis is log scale.}
\end{figure*}

%%%%%%%%%%%%%%%%%%%%%%%%%%%%%%%%%%%%%%%%%%
\subsection{HyperTree Optimizer, losses, metrics, and preprocessing}

HyperTree search repeatedly runs a sampling from 100 random architectures and then estimates 10 additional random architectures by optimizing the Expected Improvement (EI) in training loss with a predictive Sparse Gaussian Process Model (SGPM). 
These limits were chosen due to the practical tradeoff between processing time and memory utilization when evaluating the SGPM, since we found GPyOpt prediction time outstripped model evaluation time with large sample sizes.
During training we perform optimization with Stochastic Gradient Descent (SGD). We also evaluated the Adam optimizer but we found it converged to a solution less reliably. Mini-batches consist of a random example sampled at a random time step. Input to the network includes the initial image plus the image, encoded pose, and one-hot encoded action ID at the randomly chosen time step. The input gripper pose was encoded as described in Fig. \ref{fig:attempt} at that time step as an input to the network. The output of the neural network is a prediction of either the x, y, z coordinate at the goal time step encoded as $v^g_t$ (Eq. \ref{eq:translation_appendix}, \ref{eq:goal_appendix}), or the angle-axis encoded rotation $r^g_t$ (Eq. \ref{eq:rotation_appendix}, \ref{eq:goal_appendix}) at the goal time step $g$.

We initialize the network by loading the pretrained ImageNet weights when available, and otherwise weights are trained from scratch utilizing He et. al. initialization~\cite{2015heinitalization}.
During HyperTree architecture search, we evaluate each model after a single epoch of training, so we either utilize a fixed reasonable initial learning rate such as 1.0, and for longer final training runs we utilized either a triangular or exp\_range (gamma=0.999998) cyclical learning rate~\cite{2015clr} with a cycle period of 8 epochs, maximum of 2.0, and minimum of $10^{-5}$. 

Translation training is augmented with cutout~\cite{2017cutout} plus random input pose changes of up to 0.5cm and 5 degrees. Each colored block is a 5.1 cm cube. 
An example of training for the final rotation model is shown in Fig. \ref{fig:trainingplot}.

% \begin{figure}[bt!]
% \centering
% \includegraphics[width=\columnwidth]{semantic_translation_regression.pdf}
% \caption{\label{fig:trainingplot} Training progress of the best translation regression HyperTree model. Here we can see that (TODO: update figure and add brief summary)}
% \end{figure}

\subsection{HyperTree Search Heuristics}

We incorporated several heuristics to improve HyperTree search efficiency. We found that the best models would quickly make progress towards predicting goal poses, so if models did not improve beyond 1m accuracy within 300 batches, we would abort training of that model early. We also found that some generated models would stretch, but not break the limits of our hardware, leading to batches that can take up to a minute to run and a single epoch training time of several hours, so we incorporated slow model stopping where after 30 batches the average batch time took longer than 1 second we would abort the training run. In each case where the heuristic limits are triggered we return an infinite loss for that model to the Bayesian search algorithm.

\begin{figure*}[btp!]
\centering
\vspace{0.1cm}\vspace{0.1cm}
\includegraphics[width=1.0\textwidth]{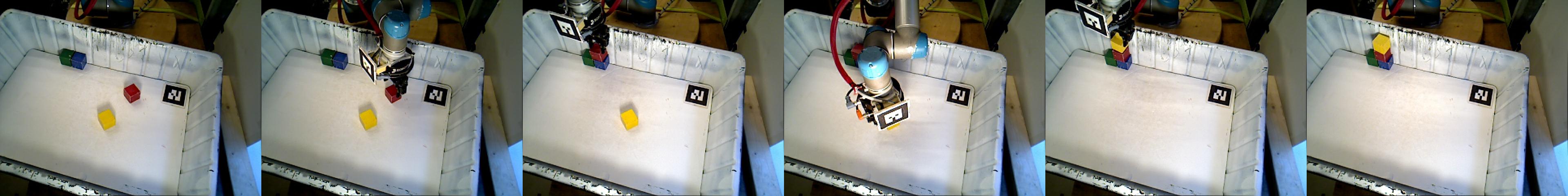}
\vspace{0.1cm}\vspace{0.1cm}
\includegraphics[width=1.0\textwidth]{2018-05-15-15-20-21_example000006_success_tiled.jpg}
\vspace{0.1cm}\vspace{0.1cm}
\includegraphics[width=1.0\textwidth]{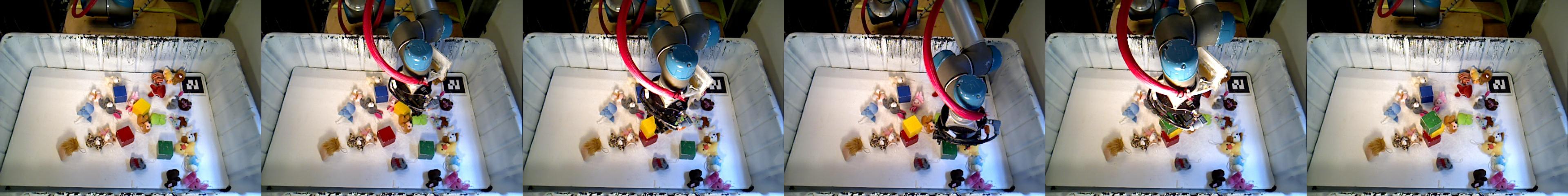}
\vspace{0.1cm}\vspace{0.1cm}
\includegraphics[width=1.0\textwidth]{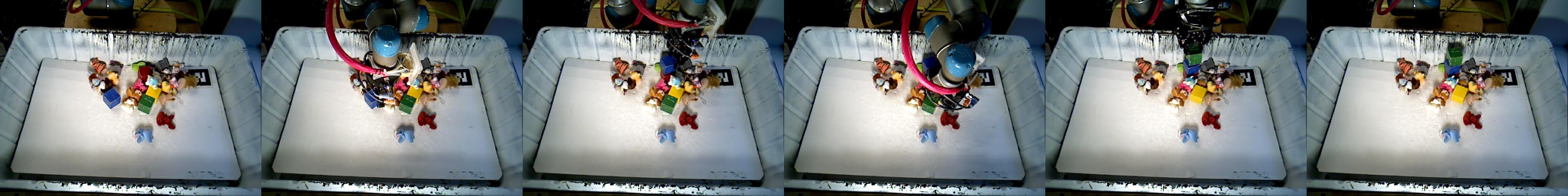}
\vspace{0.1cm}\vspace{0.1cm}
\includegraphics[width=1.0\textwidth]{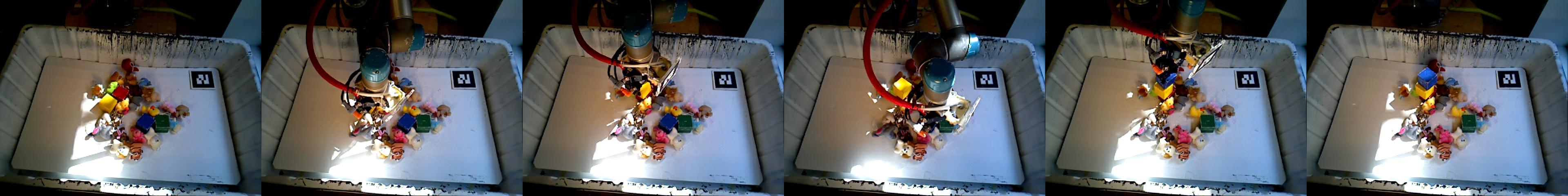}
\vspace{0.1cm}\vspace{0.1cm}
\includegraphics[width=1.0\textwidth]{2018-07-31-20-22-20_example000001_failure_tiled.jpg}

\caption{ Examples from the dataset from top to bottom with key time steps in each example from left to right. 
All rows except the bottom represent successful stacking attempts. Also see the description in Fig. \ref{fig:attempt}. 
Viewing video and other details is highly recommended, see \url{sites.google.com/site/costardataset}.
% \ref{fig:MotionNetwork}.
}
\label{fig:attempt_appendix}
\vspace{-0.4cm}
\end{figure*}
\end{document}